\documentclass[runningheads]{llncs}
\usepackage{graphicx}
\usepackage{amsmath,amssymb} 

\newcommand{\ie}{\emph{i.e}.~}

\begin{document}

\title{Contextual Object Detection with a Few Relevant Neighbors} 
\titlerunning{Contextual Object Detection with a Few Relevant Neighbors} 


\author{Ehud Barnea \and Ohad Ben-Shahar}
\authorrunning{Ehud Barnea \and Ohad Ben-Shahar} 

\institute{Dept. of Computer Science, Ben-Gurion University\\
	Beer-Sheva, Israel\\
	\email{\{barneaeh, ben-shahar\}@cs.bgu.ac.il}}

\maketitle

\begin{abstract}
A natural way to improve the detection of objects is to consider the contextual constraints imposed by the detection of additional objects in a given scene. In this work, we exploit the spatial relations between objects in order to improve detection capacity, as well as analyze various properties of the contextual object detection problem. To precisely calculate context-based probabilities of objects, we developed a model that examines the interactions between objects in an exact probabilistic setting, in contrast to previous methods that typically utilize approximations based on pairwise interactions. Such a scheme is facilitated by the realistic assumption that the existence of an object in any given location is influenced by only few informative locations in space. Based on this assumption, we suggest a method for identifying these relevant locations and integrating them into a mostly exact calculation of probability based on their raw detector responses. This scheme is shown to improve detection results and provides unique insights about the process of contextual inference for object detection. We show that it is generally difficult to learn that a particular object {\em reduces} the probability of another, and that in cases when the context and detector strongly disagree this learning becomes {\em virtually impossible} for the purposes of improving the results of an object detector. Finally, we demonstrate improved detection results through use of our approach as applied to the PASCAL VOC and COCO datasets.

\keywords{Context \and Object detection.}
\end{abstract}

\section{Introduction}

The task of object detection entails the analysis of an image for the identification of all instances of objects from predefined categories \cite{Felzenszwalb_McAllester_2008_CVPR,Girshick_etal_2014_CVPR}. While most methods employ local information, in particular the appearance of individual objects \cite{Dalal_Triggs_2005_CVPR}, the contextual relations between objects were also shown to be a valuable source of information \cite{Oramas_De-Raedt_Tuytelaars_2013_ICCV,Perko_Leonardis_2010_CVIU}. Thus, the challenge is to combine the local appearance at each image location with information regarding the other objects or detections. 

Even when focusing on how context could influence the detection of a single object, one immediately realizes that the difficulty stems from the varying number of objects that can be used as predictors, their individual power of prediction, and more importantly, their {\em combined} effect as moderated by the complex interactions between the predictors themselves. Unfortunately, however, 
most previous works that employ relations between objects and their context focused on pairwise approximations  \cite{Torralba_etal_2004_NIPS,Perko_Leonardis_2010_CVIU,Arbel_etal_2017_arxiv}, assuming that the different objects that serve as sources of contextual information do not interact among themselves. 

More current detectors based on convolutional neural networks are also able to reason about context since the receptive field of neurons grows with depth, eventually covering the entire image. However, the extent to which such a network is able to incorporate context is still not entirely understood \cite{Luo_etal_2016_NIPS}. To include more explicit contextual reasoning in the detection process, several approaches suggested to include additional layers such as bidirectional recurrent neural networks (RNNs) \cite{Bell_etal_2016_CVPR}, or attention mechanisms \cite{Li_etal_2017_TransOnMulti}. These methods have shown to improve detection results, but the types of contextual information they can encode remains unclear. Additionally, such networks are not able to reason about object relations in a manner invariant to viewpoint, requiring training data in which all meaningful relations between all groups of objects are observed from all relevant viewpoints.

\begin{figure}
	\begin{center}
		\includegraphics[width=0.5\linewidth]{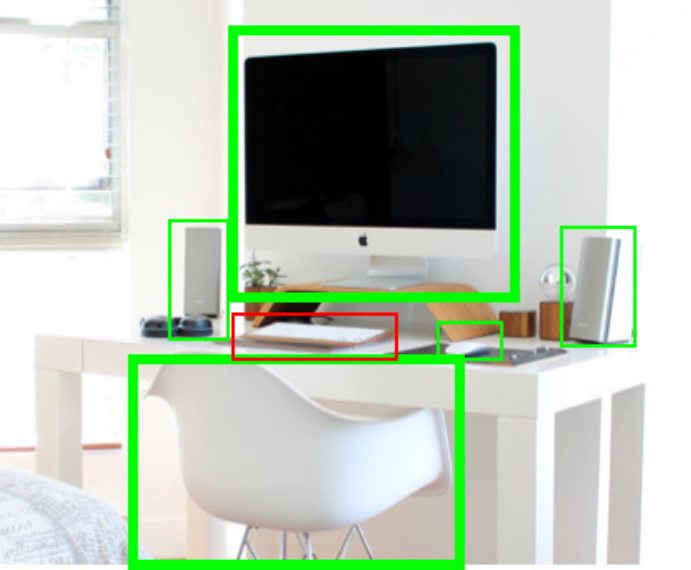}
	\end{center}
	\caption{
		An example scene with overlaid detections where thicker boxes represent higher confidence. Employing the green box detections as the context of a query red box detection can affirm (or reject) the presence of a keyboard.
		In this case, the contextual information supplied by the low confidence speakers and mouse detections may become redundant once the confidently detected chair and monitor have been accounted for. In this paper we propose a rigorous model that makes a decision for each query detection by identifying and using only its most informative detections for a more tractable decision.
	}
	\label{fig:scene}
\end{figure}

As mentioned above, some complications in addressing the full fledged contextual inference problem may emerge from attempting to model the relations between all (detected or predicted) objects to {\em all} other detections or image locations. But in reality, such comprehensive contextual relations are rarely observed or needed, as the existence of objects (or lack thereof) is correlated to just {\em few} other locations in space (and thus to the detected objects in those locations). This notion is exemplified in Figure~\ref{fig:scene}. In this paper we employ exactly this assumption to calculate a score for any given query detection. To do so, we first define the relevant context for that detection as the (few) other most informative detections, and then we use only these detections to calculate (in a closed form fashion) the probability that the query detection is indeed an object.

The suggested approach, facilitated by the decision to employ only few informative detections, provides several contributions. First, it is shown to improve the results of state-of-the-art object detectors. Second, unlike previous methods that require costly iterative training procedures, training our model is as quick and simple as just counting. Third, we represent object relations in a framework that allows to incorporate scale-invariant relations, reducing the number of needed examples, thus simplifying the training phase even further. Finally, using the derived calculation we observe various aspects and obtain novel insights related to the contextual inference of objects. In particular, we show that the effect of context is relative to the prior probability of the query object. As we show, this typically small quantity makes it difficult to infer when an object \emph{reduces} the probability of another, and it practically prohibits the {\em improvement} of detection probability when the context {\em strongly disagrees} with the raw detector result. These and further observations and insights are analyzed 
in our results.

\section{Relevant Work}
\label{sec:prev_work}

Valuable information regarding image or scene elements may be obtained by examining their context. Indeed, different kinds of context were employed for various inference problems \cite{Torralba_Sinha_2001_ICCV,Rabinovich_etal_2007_ICCV,Heitz_Koller_2008_ECCV,Hoiem_etal_2008_IJCV,Mairon_Ben-Shahar_2014_ECCV}. 
In this work we focus on employing long-range spatial interactions between objects. This problem has also gained
attention, where the standard framework is that of employing fully-connected Markov random fields (MRF) and
conditional random fields (CRF) with pairwise potentials \cite{Torralba_etal_2004_NIPS,Desai_etal_2011_IJCV,Mottaghi_etal_2014_CVPR,Galleguillos_etal_2008_CVPR}. This pairwise assumption is at the heart of the decision
process, as it entails that a decision about an object is made by employing information supplied by its neighbors but
without considering interactions {\em between} these neighbors. 
Other notable works make such assumption, where all detections affect each other in a voting scheme that is weighted by confidence \cite{Perko_Leonardis_2010_CVIU}, as a sum weighted by the probability of an object given by the detector \cite{Oramas_De-Raedt_Tuytelaars_2013_ICCV}, or a more complex voting mechanism that favors the most confident hypotheses and gives higher relevance to the object relations observed during training \cite{Oramas_De-Raedt_Tuytelaars_2014_WACV}. 
Yet another popular tool for reasoning with many elements is to employ linear classifiers \cite{Felzenszwalb_etal_2010_PAMI,Wolf_Bileschi_2006_IJCV} or more complicated set classifiers \cite{Cinbis_Sclaroff_2012_ECCV} in a pairwise interaction scheme. 

Seeking more accurate representations, schemes that incorporate higher order models have been suggested for different kinds of problems in computer vision other than contextual object detection~\cite{Kohli_Rother_2004_Higher-Order}. One such noteworthy model was suggested for objects that were not detected at all, where new detection hypotheses are generated by sampling pairwise and higher order relations using methods from topic modeling \cite{Oramas_Tuytelaars_2016_CVIU}. While this method was facilitated by context, it did not play a role in re-scoring existing or new detections. Other higher order models include neural networks that implicitly or explicitly reason about context as discussed in the Introduction. 

A fundamental aspect of our work deals with finding the most relevant set of location variables for a prediction about another location. This problem can be abstracted as finding the structure of a graph (even if only locally) and algorithms for doing so can be grouped to roughly three types \cite{Koller_Friedman_2009_ProbabilisticGraphicalModels}. {\em  Constraint-based methods} make local decisions to connect dependent variables, {\em score-based methods} penalize the entire graph according to an optimization criterion, and {\em model averaging methods} employ multiple graph structures. While our problem is better related to the constraint-based approach, most algorithms in this class seek to find the structure without considering current beliefs, a set of measures that change dramatically after observing the detections. 
To better support such cases, Chechetka and Guestrin~\cite{Chechetka_Guestrin_2010_NIPS} proposed to learn evidence-specific structures for CRFs, in which a new structure is chosen based on given evidence. This approach, however, is limited to trees. In a different approach, contextual information sources were dynamically separated to those that accept or reject each detection \cite{Yu_etal_2016_BMVC}, but in a non-probabilistic framework. To facilitate structure learning in our extended graph configurations we therefore propose a different regime based on local structure exploration for each variable during the process of belief propagation. As will be discussed, our computational process is inspired by Komodakis and Tziritas \cite{Komodakis_Tziritas_2007_PAMI} since it prioritizes variables for the message passing process according to their confidence regarding the labels they should be assigned.

\section{Suggested Approach}

The input to our algorithm is a set of detections \mbox{$\mathcal{Y}=\{Y_1,Y_2,...,Y_n\}$}, such that each detection \mbox{$Y_i=(t_i,l_i,s_i,c_i)$} comprises type $t_i$, location $l_i$, size $s_i$, and confidence $c_i$. A random variable $X_i$ is created for each detection $Y_i$, denoting the probability of having an object of type $t_i$ at location $l_i$ with size $s_i$. For the sake of brevity, in the remainder of this text we refer to $X_i$ as representing an empty location if it is indeed empty, or as a location containing an object of different type or size. 

Our goal is to calculate a new confidence for each location variable $X_i$ using detections $\mathcal{Y}$. To do so, we calculate the probability $P(X_i|\mathcal{Y})$ in a belief propagation process, where the context of $X_i$ is dynamically selected as the most informative {\em small} set $\mathcal{N}_i$ of location variables $\mathcal{X}_{j \ne i}$, which is the set of all location variables except $X_i$. 
The initial beliefs are determined according to the detector, followed by iterations in which an updated belief is calculated for each $X_i$ by identifying the best set $\mathcal{N}_i$ and employing the current belief of its variables for a decision about $X_i$.
Alas, it turns out that this calculation can be very sensitive and produce problematic results when the detector and context strongly disagree. We therefore identify these cases first, then calculate $P(X_i|\mathcal{Y})$ accordingly.

\subsection{Calculation of object probability}
\label{sec:object_prob}

\begin{figure}
	\begin{center}
		\includegraphics[width=0.3\linewidth]{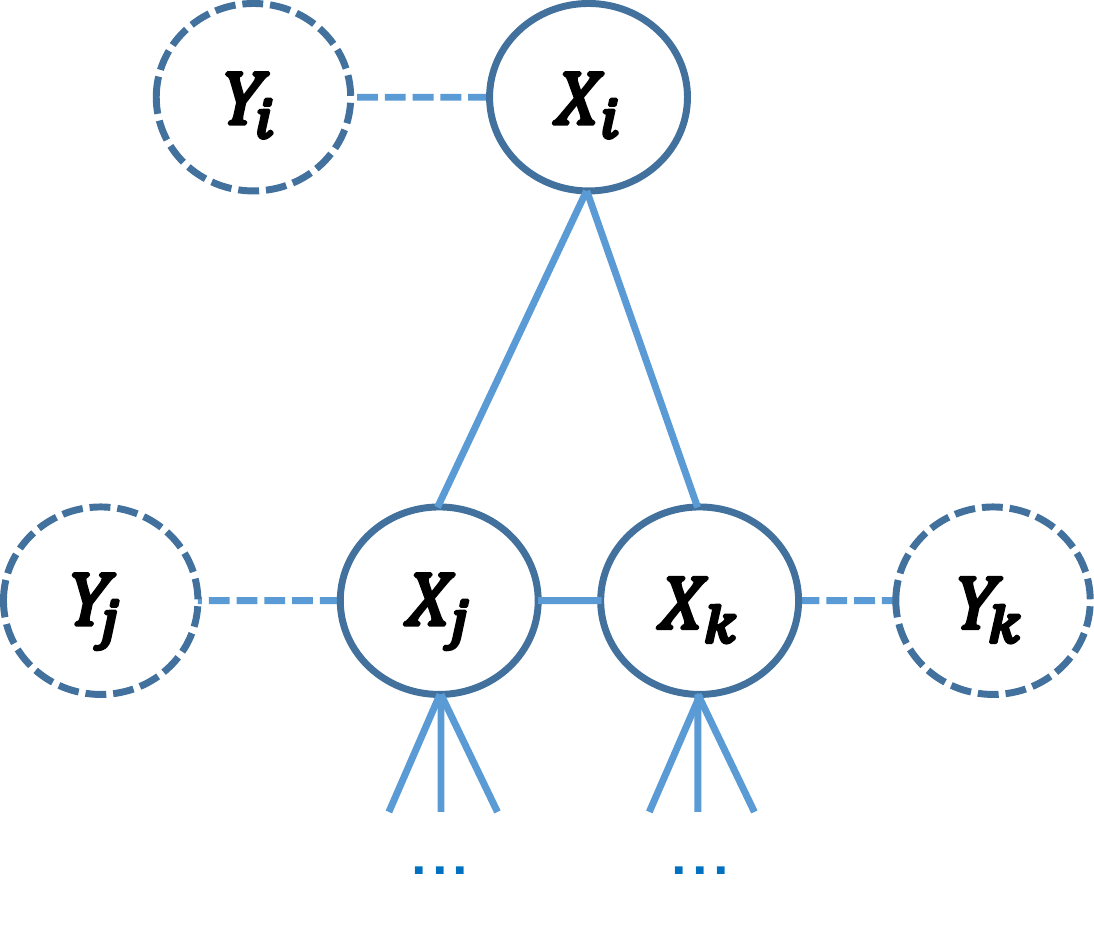}
	\end{center}
	\caption{Assumed graph structure when calculating $P(X_i|\mathcal{Y})$ in an iteration in which $\mathcal{N}_i=\{X_j,X_k\}$ is the set of most relevant neighbors for the location variable $X_i$. Each detection variable from $\mathcal{Y}$ is associated with a single location variable, the query $X_i$ and the variables in $\mathcal{N}_i$ form a clique, and the rest of the variables are connected to $X_i$ only via $\mathcal{N}_i$. We note that different graph structures are used for each $X_i$ in each iteration according to current beliefs.}
	\label{fig:graph_part}
\end{figure}

We assume that location and detection variables are connected according to the graph structure shown in Figure~\ref{fig:graph_part}. More formally, we assume that detection $Y_i$ directly depends only on the existence of an object at $X_i$, so 
\begin{align}
	\label{eq:assum1}
	P(Y_i,\mathcal{Y}_{j\ne i} | X_i)=P(Y_i | X_i)P(\mathcal{Y}_{j\ne i} | X_i) \;\; .
\end{align}
We further assume that $X_i$ directly depends only on its detection $Y_i$ and on a small set $\mathcal{N}_i$ of location variables. Therefore
\begin{align}
	\label{eq:assum2}
	P(\mathcal{Y}_{j\ne i} | X_i, \mathcal{N}_i)=P(\mathcal{Y}_{j\ne i} | \mathcal{N}_i)  \;\; .
\end{align}
We note that variables in the set $\mathcal{N}_i$ may or may not directly depend on each other. 

Employing these assumptions and the set $\mathcal{N}_i$ (identified as described in Sec. \ref{sec:src_ident}), we calculate $P(X_i|\mathcal{Y})$ in the following way:
\begin{flalign}
	\nonumber
	P(X_i|\mathcal{Y})=\sum_{\mathcal{N}_i} P(X_i,\mathcal{N}_i|\mathcal{Y})
\end{flalign}
Applying Bayes' rule we first obtain
\begin{flalign}
	\nonumber
	P(X_i|\mathcal{Y})=\sum_{\mathcal{N}_i} P(\mathcal{Y}|X_i,\mathcal{N}_i)P(X_i,\mathcal{N}_i)\frac{1}{P(\mathcal{Y})} \;.    
\end{flalign}
Employing Eq. \ref{eq:assum1} entails
{\small
	\begin{flalign}
		\nonumber
		P(X_i|\mathcal{Y})=\sum_{\mathcal{N}_i} P(Y_i|X_i,\mathcal{N}_i)P(\mathcal{Y}_{j\ne i}|X_i,\mathcal{N}_i)  P(X_i,\mathcal{N}_i)\frac{1}{P(\mathcal{Y})}   \\ 
		\nonumber
		P(X_i|\mathcal{Y})=\sum_{\mathcal{N}_i} P(Y_i|X_i)P(\mathcal{Y}_{j\ne i}|X_i,\mathcal{N}_i)  P(X_i,\mathcal{N}_i)\frac{1}{P(\mathcal{Y})} \;.    
	\end{flalign}
}
Employing Eq. \ref{eq:assum2} provides
\begin{flalign}
	\nonumber
	P(X_i|\mathcal{Y})=\frac{1}{P(\mathcal{Y})} P(Y_i|X_i) \sum_{\mathcal{N}_i} P(\mathcal{Y}_{j\ne i}|\mathcal{N}_i)  P(X_i,\mathcal{N}_i)  \;\; ,  
\end{flalign}
and applying Bayes' rule again provides
{\small
	\begin{flalign}
		\nonumber
		P(X_i|\mathcal{Y})=\frac{P(Y_i|X_i)}{P(\mathcal{Y})} \sum_{\mathcal{N}_i}    P(\mathcal{N}_i|\mathcal{Y}_{j\ne i})\frac{P(\mathcal{Y}_{j\ne i})}{P{(\mathcal{N}_i)}}     P(X_i,\mathcal{N}_i) \;\; , 
	\end{flalign}
}
which results in
{\small
	\begin{flalign}
		\label{eq:pearl}
		P(X_i|\mathcal{Y}) = \frac{P(\mathcal{Y}_{j\ne i})}{P(\mathcal{Y})}P(Y_i|X_i) \sum_{\mathcal{N}_i}    P(\mathcal{N}_i|\mathcal{Y}_{j\ne i})   P(X_i|\mathcal{N}_i)  \;\; , 
	\end{flalign}
}
an expression reminiscent of the belief propagation process suggested by Pearl \cite{Pearl_1988_Probabilistic_Reasoning}.
This is further developed by applying Bayes' rule once more:
{\small
	\begin{flalign}
		\nonumber
		P(X_i|\mathcal{Y}) =\frac{P(Y_i)P(\mathcal{Y}_{j\ne i})}{P(\mathcal{Y})}P(X_i|Y_i) \sum_{\mathcal{N}_i}    P(\mathcal{N}_i|\mathcal{Y}_{j\ne i})   \frac{P(X_i|\mathcal{N}_i)}{P(X_i)}  \;. 
	\end{flalign}
}
Finally, we denote the first term with $\xi$, which is a normalizing constant and need not be explicitly calculated. We thus obtain:
\begin{flalign}
	\label{eq:obj_prob}
	P(X_i|\mathcal{Y}) = \xi \cdot P(X_i|Y_i)  \sum_{\mathcal{N}_i}    P(\mathcal{N}_i|\mathcal{Y}_{j\ne i})   \frac{P(X_i|\mathcal{N}_i)}{P(X_i)}     
\end{flalign}
an expression we assert is more informative than Eq. \ref{eq:pearl}, as it is now apparent that the way in which the context $\mathcal{N}_i$ affects $X_i$ is relative to its prior probability $P(X_i)$. We note that instead of calculating $\xi$, we normalize the values of $P(X_i|\mathcal{Y})$ calculated with $\xi=1$, so that their sum equals to 1 as in a standard belief propagation process \cite{Pearl_1988_Probabilistic_Reasoning}.

As can be seen, apart from $P(\mathcal{N}_i|\mathcal{Y}_{j\ne i})$ this expression contains only functions over a small number of variables (for small sizes of $\mathcal{N}_i$). We therefore restrict the size of $\mathcal{N}_i$ to the maximum that still enables to properly represent and learn these functions, as employing more neighbors would require more memory and more training examples.

We are left with the term $P(\mathcal{N}_i|\mathcal{Y}_{j\ne i}) $, which is more complicated to calculate. This term is the (joint) belief of variables in $\mathcal{N}_i$ given all detections but $Y_i$, and can be seen as a weighting factor to the extent we are confident about assignments for $\mathcal{N}_i$. We therefore suggest to approximate it as the product of individual beliefs of variables in $\mathcal{N}_i$:
\begin{align}
	\label{eq:neighbors_belief}
	P(\mathcal{N}_i|\mathcal{Y}_{j\ne i})  = \Pi_{X_j \in \mathcal{N}_i} P(X_j|\mathcal{Y}_{j\ne i})  \;\;.
\end{align}
With this approximation, the representation of $P(X_i|\mathcal{Y})$ based on Eqs. \ref{eq:obj_prob} and \ref{eq:neighbors_belief} now consists of simple
functions (that are easily measured from data), and terms of the form $P(X_j|\mathcal{Y}_{j\ne i})$, which can be seen as the
messages in a standard belief propagation process. 

Finally, given detections $\mathcal{Y}$ provided by a base detector applied to an image, a new confidence is calculated for each detection $X_i$. Its most informative neighbors $\mathcal{N}_i$ are identified as explained in Sec. \ref{sec:src_ident}, and used for the calculation of the probability $P(X_i|\mathcal{Y})$ in a belief propagation process. The confidence assigned to $X_i$ is then $P(X_i=True|\mathcal{Y})$.

\subsection{Scale invariant representation}
\label{sec:vp_inv}

The term $P(X_i|\mathcal{N}_i)$ in Eq. \ref{eq:obj_prob} represents relations between several locations. Such a term must be
calculated for each set of locations, requiring to observe many examples of object groups in each location. To reduce
this complexity we make the (very reasonable) assumption that relations between objects are independent of the viewer, and suggest a representation that is invariant to different object scales. 

Similar to the spatial features employed by Cinbis and Sclaroff \cite{Cinbis_Sclaroff_2012_ECCV}, we represent the spatial relation $P(X_i|\mathcal{N}_i)$ with respect to the size of a reference object $X_j \in \mathcal{N}_i$. The relative location of $X_i$ (and any other non-reference object $X_k \in \mathcal{N}_i$) is represented as
\begin{align}
	\frac{l_i - l_j}{s_j f_j}   \;\;,
\end{align}
where $l_i$,$l_j$ are the locations of the center points of $X_i$ and $X_j$, $s_j$ is the height of $X_j$, and $f(t_j)$ is a scaling factor for $s_j$ according to $t_j$, the type of $X_j$ assigned by the base detector. Object scales are also represented relative to the reference $X_j$
\begin{align}
	log(\frac{s_i}{s_j})   \;\;.
\end{align}
Using this representation, the probability \mbox{$P(X_i|\mathcal{N}_i \setminus X_j, X_j=True)$} is measured for any assignment to $X_i$ and to the variables in $\mathcal{N}_i$ except $X_j$, which contains an object used as a reference frame. Specifically, we count the occurrences of objects of each type in each location and scale relative to reference objects of type $t_j$ that appear in training data.

A non-parametric representation of $P(X_i|\mathcal{N}_i)$ requires a value for every possible assignment of the variables. The described method for measuring $P(X_i|\mathcal{N}_i)$ requires at least one variable $X_j \in \mathcal{N}_i$ that is not empty (\ie, contains an object) with which to construct a reference frame. However, in some assignments there may be no such $X_j$.
Hence, for assignments in which all the variables in $\mathcal{N}_i$ are empty while $X_i$ is not, we simply use $X_i$ as reference:
\begin{align}
	P(X_i|\mathcal{N}_i) = P(X_k|X_i,\mathcal{N}_i \setminus X_k) \frac{P(X_i|\mathcal{N}_i \setminus X_k)}{P(X_k|\mathcal{N}_i \setminus X_k)} \;\;,
\end{align}
where $X_k$ is arbitrarily picked from $\mathcal{N}_i$. Notice how the terms of the quotient operate on one less variable.

Finally, the probability for the assignment in which all the variables are empty is calculated by subtracting the probability of the complementary event from one. 

\subsection{The implications of high probability derivative}
\label{sec:prob_deriv}

To better understand the way contextual information is combined with the detector response we revisit
Eq.~\ref{eq:obj_prob} and examine its behavior when the context is known. Thus, for some assignment to the members 
of $\mathcal{N}_i$ we assume that $P(\mathcal{N}_i|\mathcal{Y}_{j\ne i})=1$, and so Eq.~\ref{eq:obj_prob} reduces to:

\begin{flalign}
	\label{eq:combining_det_and_context}
	&P(X_i|\mathcal{Y}) = \xi \cdot P(X_i|Y_i)   \frac{P(X_i|\mathcal{N}_i)}{P(X_i)}   \;\;.  
\end{flalign}

\begin{figure}[h]
	\begin{center}
		\includegraphics[width=0.7\linewidth]{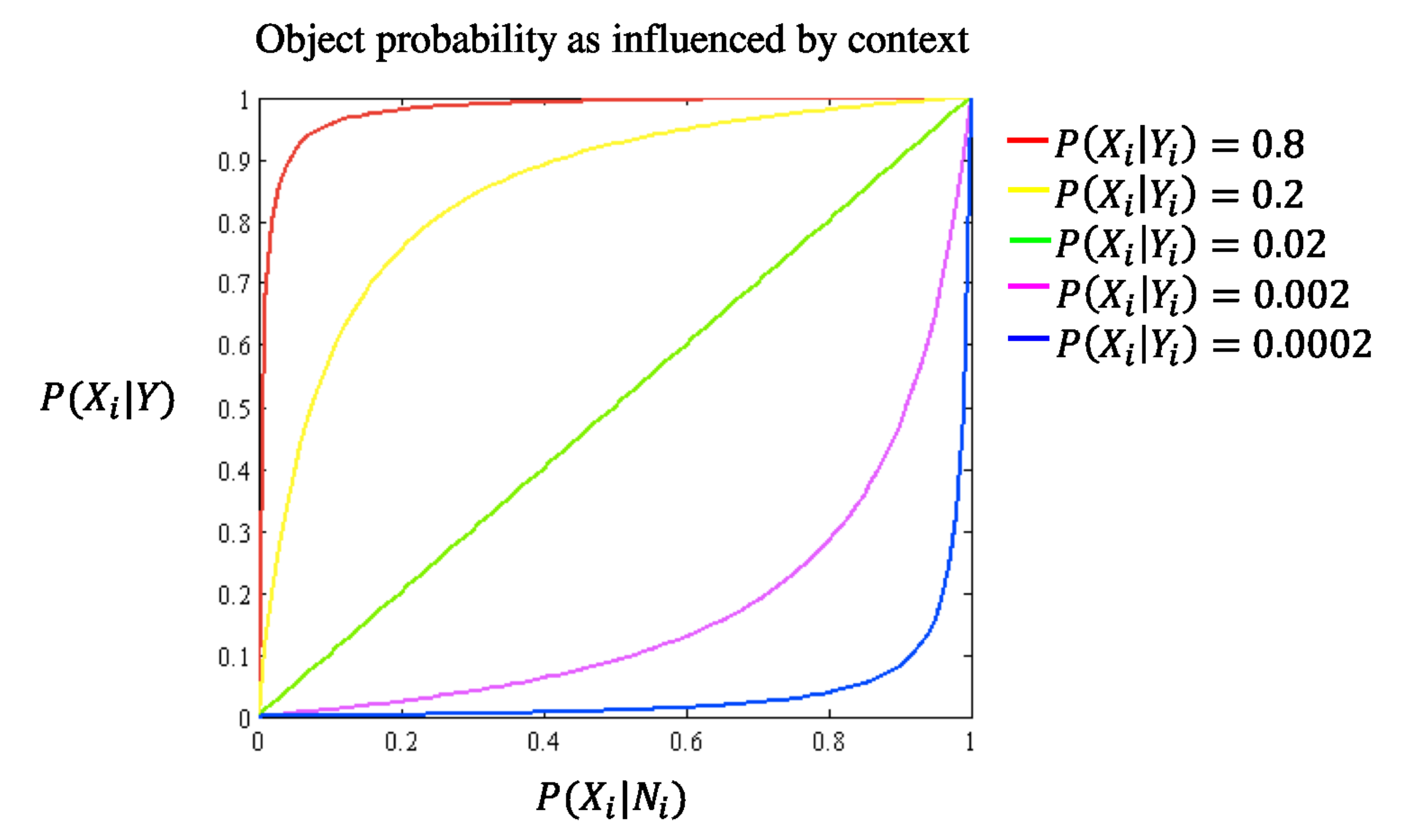}
	\end{center}
	\caption{The overall object probability $P(X_i|\mathcal{Y})$ as a function of the context-based probability $P(X_i|\mathcal{N}_i)$ for some variable assignment. Each curve represents a different value of local detector response $P(X_i|Y_i)$, for a relatively large prior probability of $P(X_i)=0.02$. This visualizes the way in which both sources of information are combined for a final decision. See the text for further analysis.}
	\label{fig:prob_deriv}
\end{figure}

A graph of Eq. \ref{eq:combining_det_and_context} for different detector responses is presented in Figure~\ref{fig:prob_deriv}. As can be seen, the addition of context strengthens a detection when the context-based probability is bigger than the prior, and weakens it when the opposite occurs. The red and blue curves, representing especially strong and weak detections respectively, \emph{exhibit large derivative regions where the detector and context strongly disagree}. It may also be the case when the detector is confident and the context is independent, \ie, $P(X_i|\mathcal{N}_i)=P(X_i)=0.02$ in the case of Fig. \ref{fig:prob_deriv}. In these cases, the overall probability greatly changes with small perturbations of the context-based probability $P(X_i|\mathcal{N}_i)$. Because this quantity is measured from data, great errors are to be expected when the number of samples does not suffice.

\begin{figure}
	\begin{center}
		\includegraphics[width=0.7\linewidth]{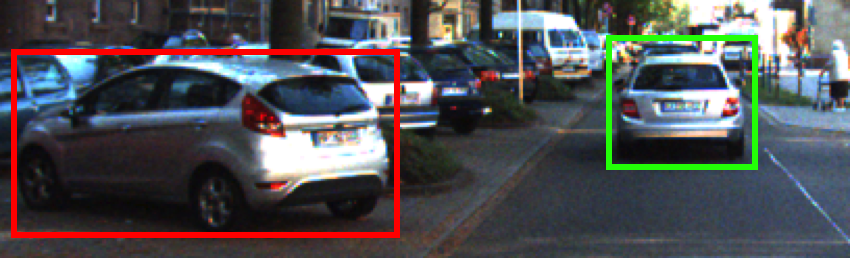}
	\end{center}
	\caption{A recurring case in which a confident detection (in the left) is assigned with a low probability. The probability of the left detection is calculated using the detector confidence and the detection in the right. While it is reasonable that the context affects the detection, it should not do so to the extent of nullifying a confident detection.}
	\label{fig:error_case}
\end{figure}

Owing to this, many cases are indeed observed where detections are incorrectly assigned with low probabilities despite a confident detector and when the context is seemingly independent. Failing to address such cases leads to poor results, as we show in Section \ref{sec:res_and_discuss}, and a specific case can be seen in Figure~\ref{fig:error_case}.
Hence, to handle such cases, for each assignment to the members of $\mathcal{N}_i$ we calculate the derivative of $P(X_i|\mathcal{Y})$ at \mbox{$x=P(X_i|\mathcal{N}_i)$} and if it exceeds a threshold we ignore the context by setting $P(X_i|\mathcal{N}_i)=P(X_i)$, essentially assuming that $X_i$ and $\mathcal{N}_i$ are independent, and thus, that the context has no effect.

\begin{figure}
	\begin{center}
		\includegraphics[width=0.5\linewidth]{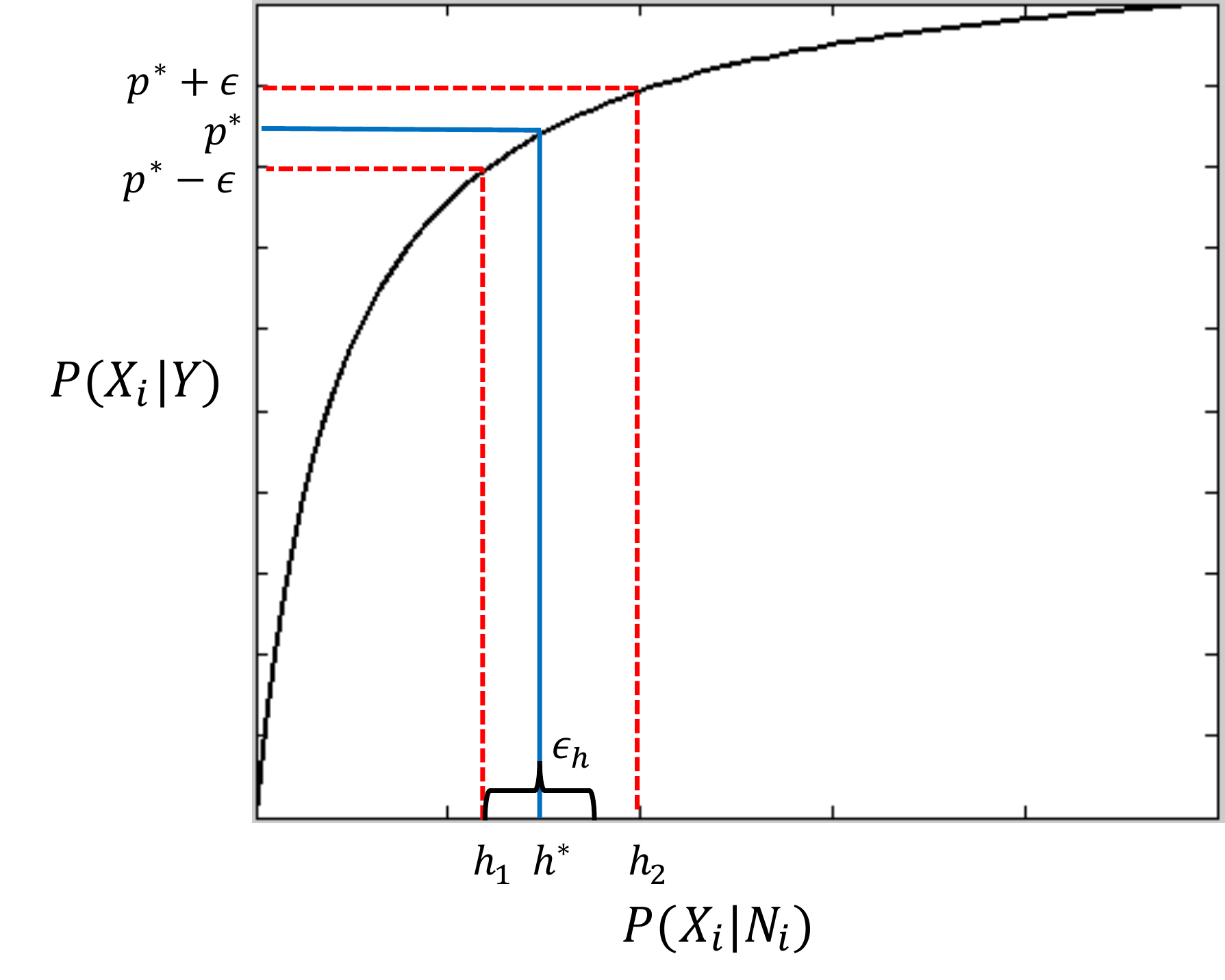}
	\end{center}
	\caption{The overall probability $P(X_i|\mathcal{Y})$ and its error in relation to $h^*$, the value of $P(X_i|\mathcal{N}_i)$ we measure from data. 
		See the text for details.}
	\label{fig:hoeffding}
\end{figure}

Another way to identify these cases is to estimate the number of samples needed to ensure a low error. Let \mbox{$p^*=P(X_i|\mathcal{Y})$} and $h^*=P(X_i|\mathcal{N}_i)$, as depicted in Figure~\ref{fig:hoeffding}. If we allow a maximal error of $\epsilon$, we require that:
\begin{align}
	\nonumber
	|p-p^*|<\epsilon  \;\;,
\end{align}
where $p$ is the value calculated using measured data. For $h$, the measured value of $P(X_i|\mathcal{N}_i)$, to provide an error that does not exceed $\epsilon$, it is required to stay within the limits of $h_1$ and $h_2$, that are the values of $P(X_i|\mathcal{N}_i)$ that correspond to $P(X_i|\mathcal{Y})=p^*-\epsilon$~ and $P(X_i|\mathcal{Y})=p^*+\epsilon$~ respectively. Thus:
\begin{align}
	\nonumber
	|h-h^*|  <  \min(|h^*-h_1|,|h^*-h_2|) = \epsilon_h \;\;,
\end{align}
where $\epsilon_h$ is the allowed measurement error for $h$.

We then employ Hoeffding's inequality \cite{Shalev-Shwartz_2014_UnderstandingML} to estimate $m$, the number of samples $Z_i$ required for $h$: 
\begin{align}
	\nonumber
	h=\frac{1}{m} \sum_{i=1}^{m}Z_i  \;\;.
\end{align}
We assume that the indicator variables $Z_i$ are sampled $i.i.d$ and note that their expectation is equal to $h^*$. Therefore, the probability of measuring $h$ with large overall error can be expressed as:
\begin{align}
	\nonumber
	P(|p-p^*|>\epsilon)<P(|h-h^*|>\epsilon_h)<2e^{-2m\epsilon_h^2}=\delta \;\;.
\end{align}
Finally, to guarantee a maximal error of $\epsilon$ with probability $\delta$, the number of required samples $m$ is:
\begin{align}
	\label{eq:num_samples}
	m>\frac{\ln (\frac{2}{\delta})}{2\epsilon_h^2} \;\;.
\end{align}

This expression enables to calculate the needed number of samples for $P(X_i|\mathcal{N}_i)$ in each case and then to employ it only when enough data is provided (even when the derivative is high). In our experiments the derivative was used to identify problematic cases, while $m$ is used for discussion in Section \ref{sec:res_and_discuss}.

\subsection{Identification of relevant detections}
\label{sec:src_ident}

The described algorithm requires to identify the most relevant set of locations $\mathcal{N}_i\in \mathcal{S}$ to use as context for $X_i$, where $\mathcal{S}$ is the set of subsets of $\mathcal{X}_{j\ne i}$. To limit the size of $\mathcal{N}_i$, we consider only subsets for which the cardinality is equal or smaller than a predefined number that is given as a parameter of the model. 

The set $\mathcal{N}_i$ was used for the assumption that $X_i$ does not depend on other locations (or their
detections) given $\mathcal{N}_i$: 
\begin{align}
	P(X_i|\mathcal{N}_i,\mathcal{Y}_{j\ne i})=P(X_i|\mathcal{N}_i) \;\;.
\end{align}

Thus, the most suitable $\mathcal{N}_i$ would be
\begin{align}
	\arg\!\min_{\mathcal{N}_i\in \mathcal{S}} \;  \sum_{X_i,\mathcal{N}_i} |P(X_i|\mathcal{N}_i,\mathcal{Y}_{j\ne i})-P(X_i|\mathcal{N}_i)|  \;\;.
\end{align}
However, this calculation requires a function over many variables $\mathcal{Y}_{j\ne i}$, which seems as complicated as calculating $P(X_i|\mathcal{Y})$. We hence suggest a different way to determine $\mathcal{N}_i$. Basically, we would like to employ the set $\mathcal{N}_i$ that would be the best predictor for $X_i$ using our current beliefs for the different variables. For this reason, we pick $\mathcal{N}_i$ as those variables that are the least independent of $X_i$ and weigh them according to the current beliefs
\begin{align}
	\label{eq:identify_n}
	\arg\!\max_{\mathcal{N}_i\in \mathcal{S}}  \;  \sum_{X_i,\mathcal{N}_i} |P(X_i|\mathcal{N}_i)-P(X_i)| \cdot P(\mathcal{N}_i|\mathcal{Y}_{j\ne i}) \;\;,
\end{align}
where $P(\mathcal{N}_i|\mathcal{Y}_{j\ne i})$ is calculated as in Eq. \ref{eq:neighbors_belief}.

\section{Results and Discussion}
\label{sec:res_and_discuss}

We evaluate the proposed approach using the PASCAL VOC 2007 dataset where initial detections are provided by the Fast R-CNN detector \cite{girshick_2015_ICCV}. Training and validation set objects are used to measure the probability of an object given its context $P(X_i|\mathcal{N}_i)$ as described in Sec. \ref{sec:vp_inv}. For the probability of an object given its detection $P(X_i|Y_i)$ we simply use the confidence $c_i$ provided by the base detector. To determine $P(X_i)$, we assume that the prior probability of an object is fixed regardless of image location and size, but depends on the object type. The value of $P(X_i)$ for each type is found by an exhaustive search to maximize the method's average precision (AP) on the training and validation set. 

\begin{figure}[h]
	\begin{center}
		\begin{tabular}{ccc}
			\includegraphics[width=0.28\linewidth]{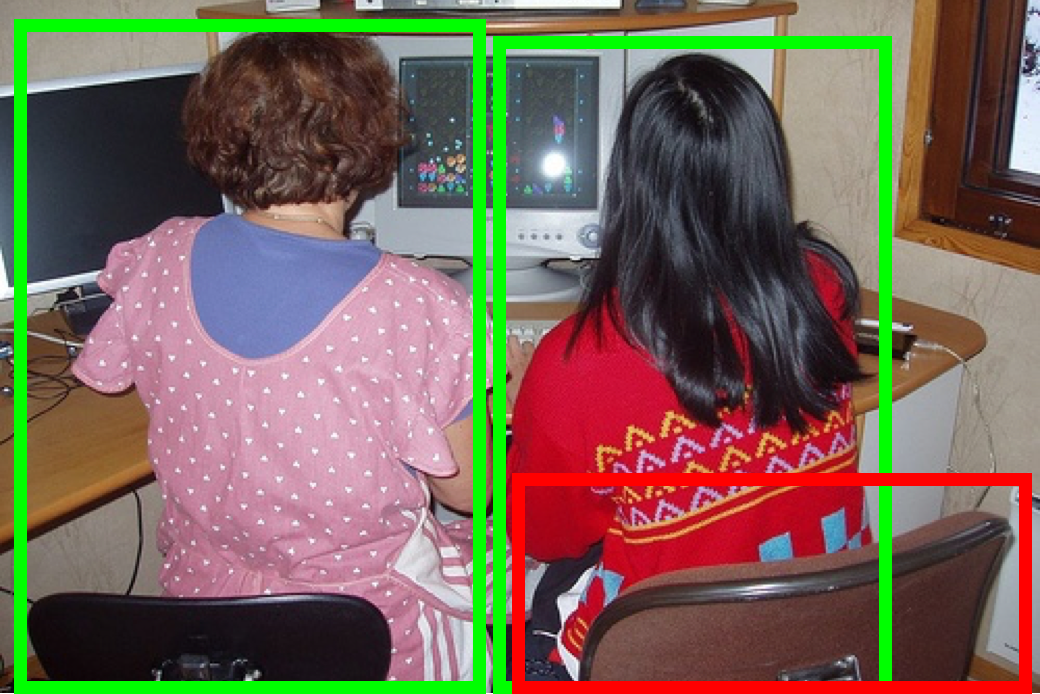} &
			\includegraphics[width=0.28\linewidth]{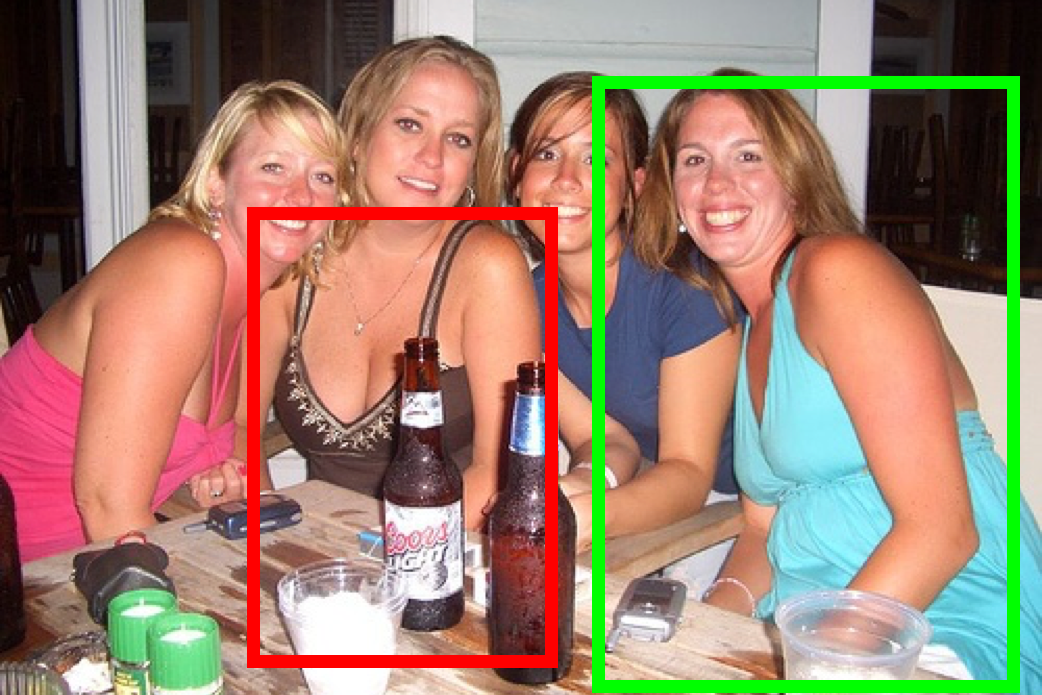} &
			\includegraphics[width=0.28\linewidth]{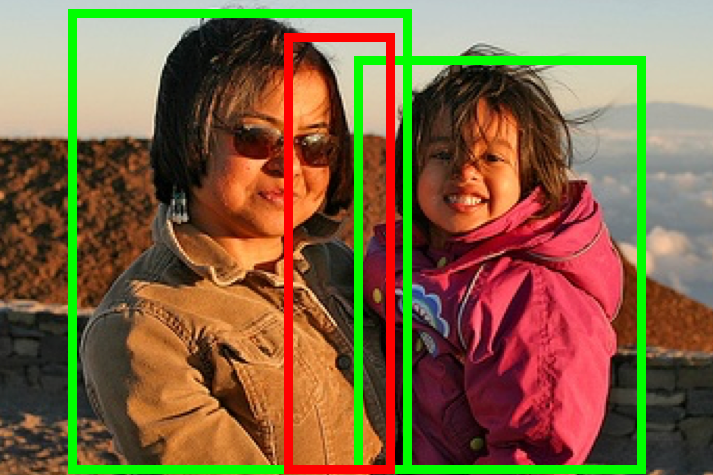} \\
			\includegraphics[width=0.28\linewidth]{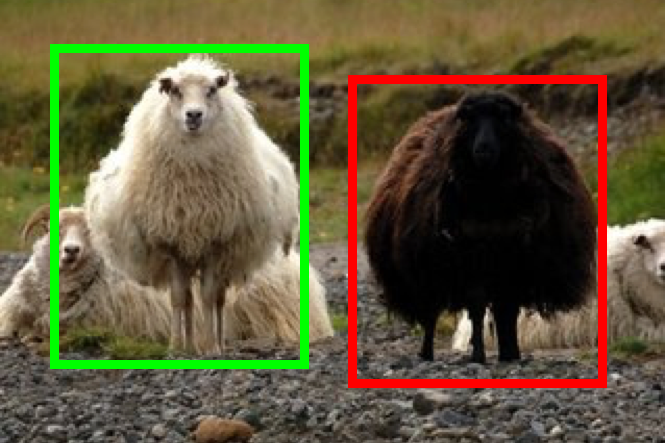} &
			\includegraphics[width=0.28\linewidth]{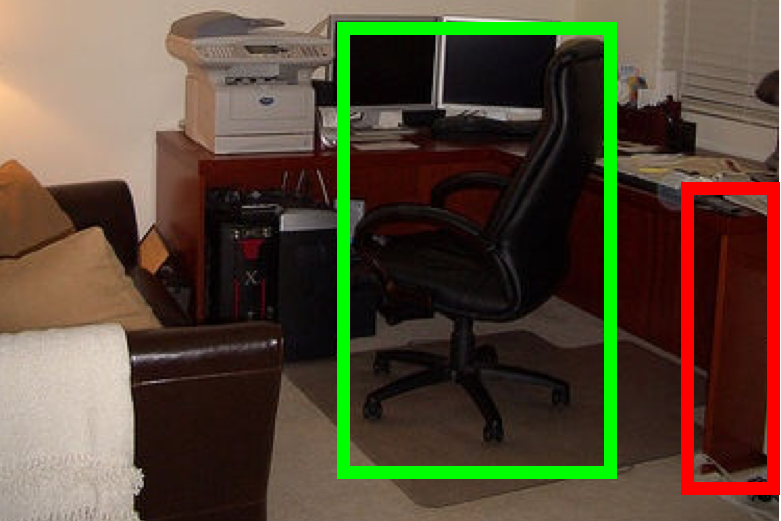} &
			\includegraphics[width=0.28\linewidth]{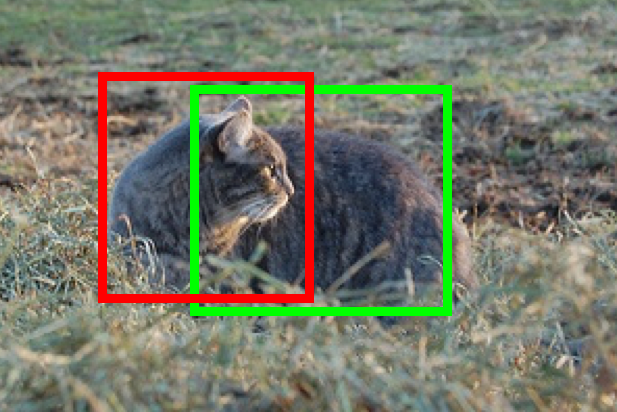} \\
			(a) & (b) & (c)
		\end{tabular}
	\end{center}
	\caption{Example detections (red) with their most informative neighbors (green). The confidence of correct chair/sheep detections in (a) increased following the use of our context model, and the confidence of the incorrect bottle (top) and tv/monitor (bottom) detections in (b) decreased. The hard to detect chair/sheep in (a) gain a boost in confidence due to the favorable context in which they appear. Similarly, The false detections in (b) are weakened due to their unusual location and size in relation to the other objects that serve as their context. Detections in (c) represent recurring mistakes due to localization errors, in which an increased confidence was provided for a false person detection (top) and a false cat detection (bottom). In both cases, object parts that were mistaken for full fledged objects are strengthened due to the use of context.
	}
	\label{fig:example_results}
\end{figure}

The results of our approach are summarized in Table~\ref{tab:results}, and specific examples including the detections identified as most informative can be seen in Figure~\ref{fig:example_results}. Included in the table are the base Fast R-CNN detector, our proposed model with two most informative detections and {\em no} treatment for large derivatives (dFNM), our model with randomly selected context (rFNM), and finally, our contextual inference model {\em with} treatment for large derivatives and two most informative detections (FNM). We note that it is also possible for a single detection to be identified as the most informative, and that employing two detections to reason about a third constitutes a triple-wise model.
In all variants of our computational model the results are reported after a single iteration of belief propagation, as usually just two or three iterations were needed for convergence. 

\begin{table}
	\begin{center}
		\resizebox{\textwidth}{!}{\begin{tabular}{ l | *{20}{c} | c }
				
				Method & aero      & bike      & bird      & boat      & bottle     & bus        & car        & cat        & chair      & cow        & table      & dog        & horse      & mbike      & persn     & plant      & sheep      & sofa       & train      & tv         & mAP \\ 
				\hline
				FRCNN & 72.15 & 72.51 & 61.19 & 42.98 & 28.67 & \textbf{69.89} & 70.67 & 81.14 & 36.15 & 69.61 & 55.90 & 77.31 & 76.69 & \textbf{69.01} & 64.55 & 28.24 & 57.35 & 59.97 & 75.58 & 62.58 & 61.61  \\ 
				\hline
				dFNM & 72.08 & 69.82 & \textbf{61.93} & 41.79 & 25.12 & 67.73 & 69.73 & 80.84 & 39.00 & 69.22 & 54.30 & 74.91 & 76.38 & 66.41 & 62.83 & 28.20 & 59.33 & 57.46 & 74.94 & 59.30 & 60.57  \\ 
				rFNM & 72.11 & 72.57 & 61.21 & \textbf{43.20} & 28.56 & 69.69 & 70.73 & \textbf{81.38} & 33.74 & 69.63 & 55.47 & 77.29 & 76.77 & 68.86 & 64.56 & 28.46 & 57.44 & \textbf{60.13} & 76.10 & 62.71 & 61.53 \\
				FNM & \textbf{72.29} & \textbf{72.63} & 61.68 & 43.15 & \textbf{29.64} & 69.81 & \textbf{70.76} & 81.36 & \textbf{39.00} & \textbf{69.73} & \textbf{56.01} & \textbf{77.63} & \textbf{76.78} & 68.93 & \textbf{64.85} & \textbf{28.90} & \textbf{59.46} & 60.09 & \textbf{76.26} & \textbf{63.58} & \textbf{62.13}  \\ 
		\end{tabular}}
	\end{center}
	\caption{Detection results in average precision (AP) of the base detector Fast R-CNN (FRCNN) and different variants of our model on all objects in the PASCAL VOC 2007 test set. Employing our model without treatment for large derivatives (dFNM) hurts detection results, and when large derivatives are treated (FNM) our model provides improved detection results. This improvement is not observed with randomly selected context (rFNM). The most significant increase is obtained for object categories such as chairs, sheep, televisions, and bottles, which benefit more from the use of context.}
	\label{tab:results}
\end{table}

As can be seen, our suggested approach (FNM) improves detection results and greater improvement is observed for the detection of chairs, sheep, televisions and bottles. The model in which large derivative regions are not handled (dFNM) is worse than the detector alone, which fits the analysis performed in Sec \ref{sec:prob_deriv}. 

For additional comparison, we test the ability of our model to improve detection results of the Faster R-CNN detector \cite{ren_etal_2015_NIPS} over the COCO dataset \cite{lin_etal_2014_ECCV}. In this case the suggested model (FNM) improved the mAP from 66.5 to 66.9 where the largest improvement of 2.6 was observed for sheep. We also compare our results to the ION contextual detector \cite{Bell_etal_2016_CVPR} that employs the same base detector and dataset (PASCAL 2007) and reports the results provided by the added context layers (but without additional components that are unrelated to context). In a nutshell, both ION and our approach provide a comparable improvement, where the mAP of the base detector was increased by 0.98 and 0.52, respectively, and our model provided a larger improvement than ION on 8 out of 20 object categories using a \emph{significantly simpler and quicker} model. Moreover, please recall that our computational approach can also be applied to the results provided by networks with context such as ION for an additional boost.

\subsection{Impracticality of learning certain properties}

The graph of Eq. \ref{eq:combining_det_and_context} in Fig.~\ref{fig:prob_deriv} sheds light on the way information supplied by the local detector is combined with the context. As can be seen, the contextual information can either increase or decrease the probability of an object, where different probabilities of the local detector response affect the rate of change.

Using Eq. \ref{eq:num_samples}, we examine the difficulty of learning different properties with regard to the number
of samples needed to stay within the limits of an allowed error. In this section we show the impracticality of
learning certain properties even under modest error requirements. We first examine the requirements for learning relations that decrease the probability of objects without necessarily
seeking to improve detections. As a test case, we set $P(X_i)=0.02$, which is a relatively high prior probability for location $X_i$ to contain an object. So, to decide with high certainty ($\delta=0.1$) whether an assignment to the
members of $\mathcal{N}_i$ reduces the probability of $X_i$, that is $P(X_i|\mathcal{N}_i)<P(X_i)$, an $\epsilon_h$ error of at most $0.02$ is required. According to Eq. \ref{eq:num_samples}, this requires at least 3745 samples of that relation (or even more for the average prior probability).

Seeking to improve detection results, let us examine one test case in which the context-based probability $P(X_i|\mathcal{N}_i)$ is half of the prior probability $P(X_i)$:
\begin{align}
	\nonumber
	P(X_i|\mathcal{N}_i)=\frac{P(X_i)}{2}=0.01  \;\;.
\end{align}
For confident detections $P(X_i|Y_i)=0.8$, the overall probability of $X_i$ to contain an object in this case is \mbox{$P(X_i|\mathcal{Y})=0.6644$}. If we require a modest overall error of at most $\epsilon=0.1$, then according to the construction
in Section \ref{sec:prob_deriv}, a measurement error less than $\epsilon_h=0.0034$ is required. In this case, for a
high certainty ($\delta=0.1$) the number of required samples is 127,095. Similarly, for more accurate results
($\epsilon=0.05$), as much as 400,048 samples are required. Of course, in more extreme cases many more samples would be required. The need to collect datasets that large renders such relations impractical to learn by observing object occurrences.

\begin{figure}
	\begin{center}
		\begin{tabular}{cc}
			\includegraphics[width=0.3\linewidth]{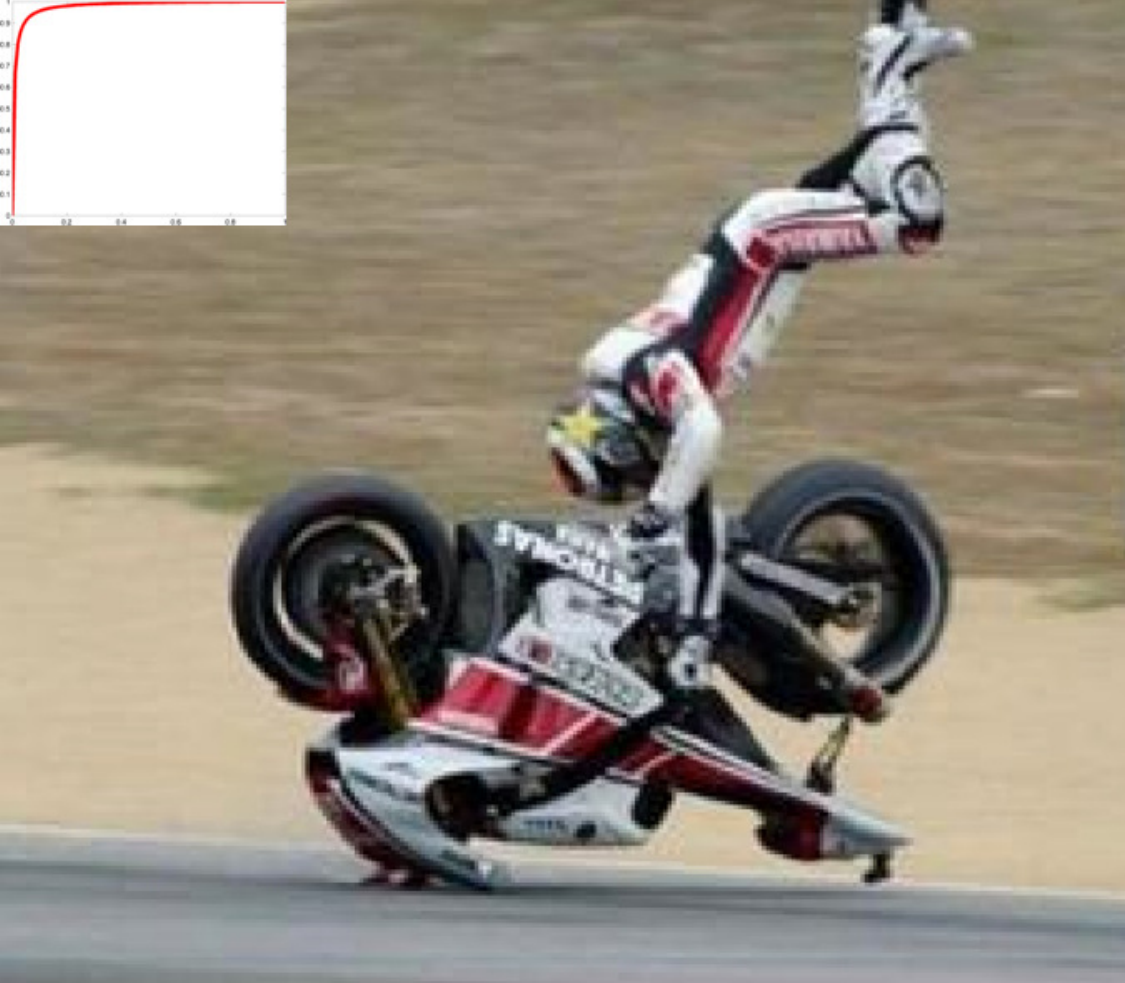} &
			\includegraphics[width=0.3\linewidth]{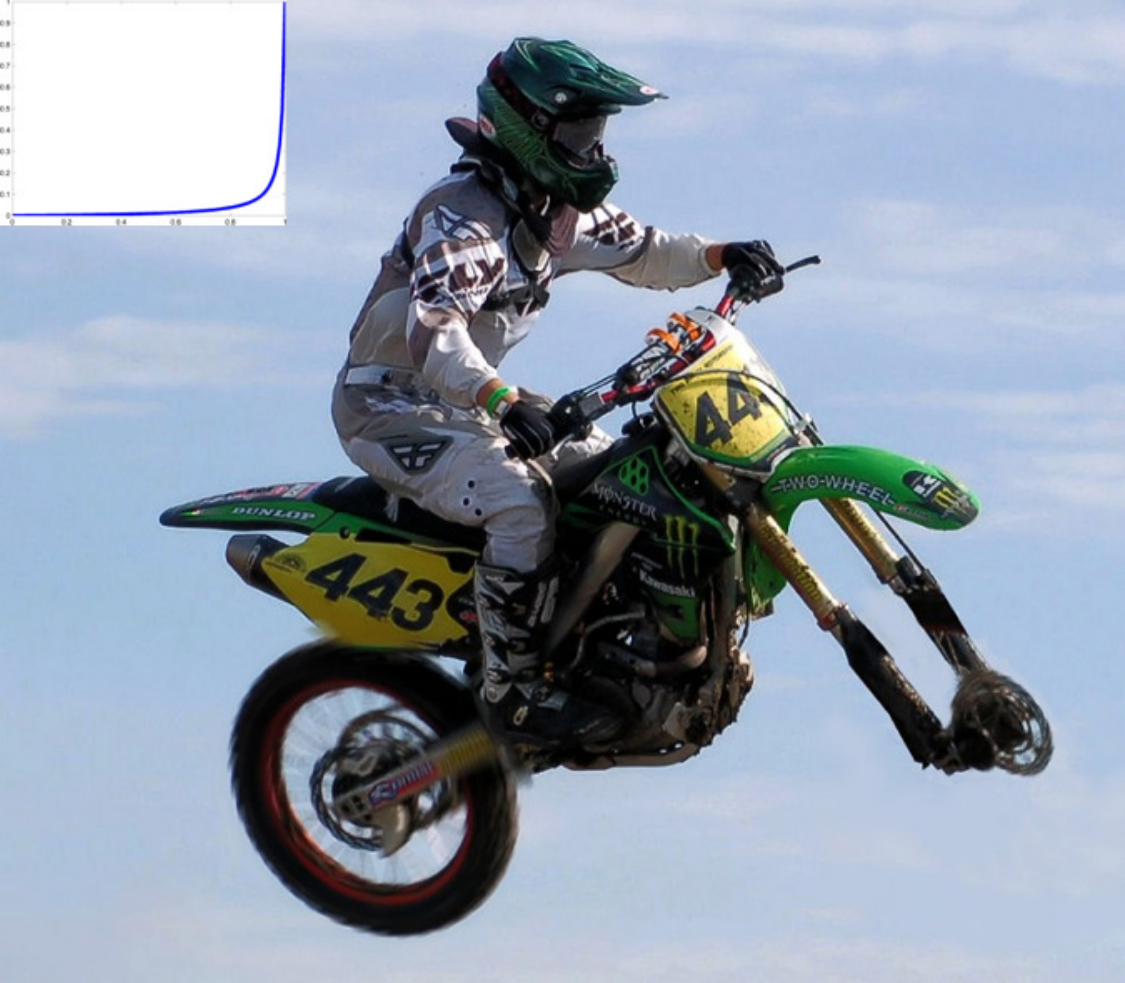} \\
			(a) & (b)
		\end{tabular}
	\end{center}
	\caption{Examples of disagreement between the context and the detector (based on local appearance) where in both cases the human visual system seems to resolve 
		this conflict by ignoring the context. The insets show the corresponding situation of the conditional probability as discussed in Fig.~\ref{fig:prob_deriv}.
		{\bf (a)} Most observers do not have a problem to immediately spot the rider despite its improbable configuration relative to the motorbike.
		{\bf (b)} When looking directly at the expected position of the front wheel, most observers report its absence even though motorbikes are mostly observed to have two wheels.
	} 
	\label{fig:human}
\end{figure}

The suggestion here was to handle the cases when the detector and context strongly disagree by assuming independence from context \mbox{$P(X_i|\mathcal{N}_i)=P(X_i)$} according to the derivative. This decision to ignore the context (instead of the detector) corresponds to similar processes we observe in the human visual system, as exemplified in Figure~\ref{fig:human}. And yet, it is important to note that there are indeed cases in which our contextual computation successfully reduces the probability of an object. Two such cases can be seen in Figure~\ref{fig:example_results}b.

Also important is the rate of change of the graph in Fig.~\ref{fig:prob_deriv} for different detector responses $P(X_i|Y_i)$ depicted by the differently colored curves. The central curve, calculated for $P(X_i|Y_i)=0.02$, behaves as a straight line. The red and yellow curves behave similarly to the blue and magenta ones. However, the scale of the latter curves is significantly smaller, reducing the high derivative cases to those in which the detector is extremely confident that an object is not present. From this we conclude that it is generally simpler to learn relations that increase the probability of an object in comparison to those that decrease its probability.




\section{Conclusions}

The problem of including context in object detection is important but difficult, as decision over many locations is needed. We have suggested to employ only a small number of locations for a more
accurate decision, and presented a method for the identification of the most informative set of locations, 
and a formulation that employs it to infer the probability of objects at different locations in a probabilistic fashion. Key benefits of our computational approach is how it facilitates
better understanding of certain aspects of the problem, and in particular it allowed to conclude that it is impractical to infer relations that decrease or increase 
the probability of detections when the detector and the context strongly disagree, or that in general it is more difficult to infer 
conditions that reduce the probability of an object rather than relations that increase it. Finally, we have demonstrated how our approach improves detection results using a model that is quick and simple to train and to employ for context-based inference.

\section*{Acknowledgments}
This research was supported in part by Israel Ministry of Science, Technology and Space  (MOST Grant 54178). We also thank the Frankel Fund and the Helmsley Charitable Trust through the ABC Robotics Initiative, both at Ben-Gurion University of the Negev, for their generous support.

\bibliographystyle{splncs04}
\bibliography{journal_names,all,ehud_cvpr2018}

\end{document}